\documentclass[runningheads]{llncs}
\usepackage[utf8]{inputenc}	
\usepackage[T2A]{fontenc}	
\usepackage[english]{babel}
\usepackage{amsfonts}
\usepackage{amsmath}
\usepackage{comment}
\usepackage{graphicx}
\usepackage{multirow}
\usepackage{enumerate}
\usepackage{paralist}
\usepackage{multicol}
\usepackage{tabularx}
\usepackage{diagbox}

\begin{document}



\title{Self-Attentive Model for Headline Generation}


 \author{Daniil Gavrilov \and Pavel Kalaidin \and Valentin Malykh}
 \institute{VK, \\
 191023, Nevsky ave., 28, Saint-Petersburg, Russia\\
\email{\{firstname.lastname\}@vk.com}}



\maketitle

\begin{abstract}
Headline generation is a special type of text summarization task. While the amount of available training data for this task is almost unlimited, it still remains challenging, as learning to generate headlines for news articles implies that the model has strong reasoning about natural language. To overcome this issue, we applied recent Universal Transformer architecture paired with byte-pair encoding technique and achieved new state-of-the-art results on the New York Times Annotated corpus with ROUGE-L F1-score 24.84 and ROUGE-2 F1-score 13.48. We also present the new RIA corpus and reach ROUGE-L F1-score 36.81 and ROUGE-2 F1-score 22.15 on it.

\end{abstract}







\section{Introduction}
Headline writing style has broader applications than those used purely within the journalism community. So-called naming is one of the arts of journalism. Just as natural language processing techniques help people with tasks such as incoming message classification (see \cite{ULMFiT} or \cite{BagOfTricks}), the naming problem could also be solved using modern machine learning and, in particular, deep learning techniques. In the field of machine learning, the naming problem is formulated as headline generation, i.e. given the text it is needed to generate a title.

Headline generation can also be seen as a special type of text summarization. The aim of summarization is to produce a shorter version of the text that captures the main idea of the source version.
We focus on abstractive summarization when the summary is generated on the fly, conditioned on the source sentence, possibly containing novel words not used in the original text.

The downside of traditional summarization is that finding a source of summaries for a large number of texts is rather costly. The advantage of headline generation over the traditional approach is that we have an endless supply of news articles since they are available in every major language and almost always have a title.

This task could be considered language-independent due to the absence of the necessity of native speakers for markup and/or model development.

While the task of learning to generate article headlines may seem to be easier than  generating full summaries, it still requires that the learning algorithm be able to catch structure dependencies in natural language and therefore could be an interesting benchmark for testing various approaches.

In this paper, we present a new approach to headline generation based on Universal Transformer architecture which explicitly learns non-local representations of the text and seems to be necessary to train summarization model. We also present the test results of our model on the New York Times Annotated corpus and the RIA corpus.


\section{Related Work}





Rush et al. \cite{rush2015sum} were the first to apply an attention mechanism to abstractive text summarization.

In the recent work of Hayashi \cite{hayashi2018headline}, an encoder-decoder approach was presented, where the first sentence was reformulated to a headline. Our Encoder-Decoder baseline (see section \ref{sec:baselines}) follows their setup. 

The related approach was presented in \cite{putra2018experiment}, where the approach of the first sentence was expanded with a so-called topic sentence. The topic sentence is chosen to be the first sentence containing the most important information from a news article (so called 5W1H information, where 5W1H stands for who, what, where, when, why, how). Our Encoder-Decoder baseline could be considered to implement their approach in OF (trained On First sentence) setup.

Tan et al. in \cite{tan2017neural} present an encoder-decoder approach based on a pregenerated summary of the article. The summary is generated using a statistical summarization approach. The authors mention that the first sentence approach is not enough for New York Times corpora, but they only use a summary for their approach instead of the whole text, thus relying on external tools of summarization.

\def\x{\boldsymbol{x}}
\def\h{\boldsymbol{h}}
\def\bc{\boldsymbol{c}}
\def\f{\boldsymbol{f}}
\def\b{\boldsymbol{b}}
\def\bu{\boldsymbol{u}}
\def\br{\boldsymbol{r}}
\def\bv{\boldsymbol{\upsilon}}


\section{Background}

Consider that we have dataset $D = \{(title_i, full text_i)\}_i^N$ of news articles and their titles. An approach for learning summarization is to define a conditional probability $P(y_t|\{y_1, ..., y_{t-1}\}, X, \theta)$ of some token $y_t \in V$ at time step $t \in \mathbb{N}$, with respect to article text $X = \{x_1, ..., x_N\}$ ($x_i \in V$ too) and previous tokens of the title $\{y_1, ..., y_{t-1}\}$, parameterized by a neural network with parameters $\theta$.

Then model parameters are found as $\theta_{MLE} = argmax_{\theta} \prod_i^N P(Y_i|X_i, \theta)$

 We can then apply two methods for finding the most probable sentence under trained model: \textit{greedy}, decoding token-by-token by finding the most probable token at each time step, and \textit{beam-search}, where we find the top-k most probable tokens at each step. The latter method yields better results though it is more computationally expensive.



Sutskever et al. \cite{sutskever2014seq2seq} proposed a model that defines $P(y_t|\{y_1, ..., y_{t-1}\}, X, \theta)$ by propagating initial sequence $X$ through a Recurrent Neural Network (RNN). Then last hidden state of RNN is used as context vector $c$ and is then passed to the second RNN with $y_1, ..., y_{t-1}$ to obtain distribution over $y_t$.

RNNs have a commonly known flaw. They rapidly forget earlier timesteps, e.g. see \cite{overcoming_catastrophic_forgetting}. To mitigate this issue, attention \cite{bahdanau2014neural}  was introduced to the Encoder-Decoder architecture. The attention mechanism makes a model able to obtain a new context vector at every decoding iteration from different parts of an encoded sequence.
It helps capture all the relevant information from the input sequence, removing the bottleneck of the fixed size hidden vector of the decoder’s RNN.

\section{Our Approach}

\subsection{Universal Transformer}
While RNNs could be easily used to define the Encoder-Decoder model, learning the recurrent model is very expensive from a computation perspective. The other drawback is that they use only local information while omitting a sequence of hidden states $H = \{h_1, ..., h_N\}$. I.e. any two vectors from hidden state $h_i$ and $h_j$ are connected with $j - i$ RNN computations that makes it hard to catch all the dependencies in them due to limited capacity. 
To train a rich model that would learn complex text structure, we have to define a model that relies on non-local dependencies in the data.

In this work, we adopt the Universal Transformer model architecture \cite{Dehghani2018universal}, which is a modified version of Transformer \cite{vaswani2017attention}.
This approach has several benefits over RNNs. First of all, it could be trained in parallel. Furthermore, all input vectors are connected to every other via the attention mechanism. It implies that Transformer architecture learns non-local dependencies between tokens regardless of the distance between them, and thus it is able to learn a more complex representation of the text in the article, which proves to be necessary to effectively solve the task of summarization. Also, unlike \cite{hayashi2018headline,tan2017neural}, our model is trained end-to-end using the text and title of each news article.

\subsection{Byte Pair Encoding}
We also adopt byte-pair encoding (BPE), introduced by Sennrich for the machine translation task in \cite{Sennrich2015bpe}. BPE is a data compression technique where often encountered pairs of bytes are replaced by additional extra-alphabet symbols. In the case of texts, like in the machine translation field, the most frequent words are kept in the vocabulary, while less frequent words are replaced by a sequence  of (typically two) tokens. E.g., for morphologically rich languages, the word endings could be detached since each word form is definitely less frequent than its stem. BPE encoding allows us to represent all words, including the ones unseen during training, with a fixed vocabulary.

\section{Experiments}
In our experiments, we consider two corpora: one in Russian and another in English. It is important to mention that we have not done any additional preprocessing other than lower casing, unlike other approaches \cite{hayashi2018headline,putra2018experiment}. We apply BPE encoding, which allows us to avoid usage of the $<UNK>$ token for out-of-vocabulary words. For our experiments, we withheld 20,000 random articles to form the test set. We have repeated our experiments 5 times with different random seeds and report mean values.

\subsubsection{English Dataset} We use the New York Times Annotated Corpus (NYT) as presented by the Linguistic Data Consortium in \cite{nyt}. This dataset contains 1.8 million news articles from the New York Times news agency, written between the years 1987 and 2006. For our experiments, we filtered out news articles containing titles shorter than 3 words or longer than 15 words. We also filtered articles with a body text shorter than 20 words or longer than 2000 words. In addition, we skipped obituaries in the dataset. After filtering, we had 1444919 news available to us with a mean title length of 7.9 words and mean text length of 707.6 words.

\subsubsection{Russian Dataset} 
Russian news agency ``Rossiya Segodnya'' provided us with a dataset (RIA) for research purposes\footnote{The dataset is available at \url{https://vk.cc/8W0l5P}}. 
It contains news documents from January, 2010 to December, 2014.
In total, there are 1003869 news articles in the provided corpus with a mean title length 9.5 words and mean text length of 315.6 words. 

\section{Experiments}
\subsection{Baseline models} 
\label{sec:baselines}
\subsubsection{First Sentence} This model takes the first sentence of an article and uses it as its hypothesis for an article headline. This is a strong baseline for generating headlines from news articles.

\subsubsection{Encoder-Decoder} 
Following \cite{putra2018experiment}, we use the encoder-decoder architecture on the first sentence of an article. The model itself is already described at recent works section as Seq-To-Seq with RNNs of Sutskever et al. \cite{sutskever2014seq2seq}. For this approach, we use the same preprocessing as we did for our model, including byte pair encoding. 


\subsection{Training}
For both datasets, NYT and RIA, we used the same set of hyper-parameters for the models, namely
4 layers in the encoder and decoder with 8 heads of attention. In addition, we added a Dropout of $p=0.3$ before applying Layer Normalization \cite{Jummy:2016:arXiv}.

The models were trained with the Adam optimizer using a scaled learning rate, as proposed by the authors of the original Transformer with the number of warmout steps equal to $4000$ in both cases and $\beta = (0.9, 0.98)$. Both models were trained until convergence.

We trained the BPE tokenizator separately on the datasets. NYT data was tokenized with a vocabulary size of active tokens equal to $40000$, while RIA data was tokenized using $50000$ token vocabulary. In addition, we have limited length of the documents with 3000 BPE tokens and 2000 BPE tokens for RIA and NYT datasets respectively. Any exceeding tokens were omitted.  
word2vec \cite{mikolov2013} embeddings were trained on each dataset with the size of each embedding equal to $512$. 
For headline generation, we adopted beam-search size of $10$.

\section{Results}
\begin{table}[!tbph]
\begin{center}
 \begin{tabular}{|l| c c c c c c|} 
 \hline
  Model & R-1-f & R-1-r & R-2-f & R-2-r & R-L-f & R-L-r\\ 
 \hline
  \multicolumn{1}{|c}{} &
 \multicolumn{6}{c|}{New York Times}\\\hline
 First Sentence & 11.64 & \textbf{34.67} & 2.28 & 7.43 & 7.19 & \textbf{31.39}\\
 Encoder-Decoder & 23.02 & 21.90 & 11.84 & 11.44 & 21.23 & 21.31 \\
 summ-hieratt \cite{tan2017neural} & - & 29.60 & - & 8.17 & - & 26.05\\
 
 Universal Transformer w/ smoothing (ours) & 25.60 & 23.90 & 12.92 & 12.42 & 23.66 & 25.27\\ [1ex] 
 
 Universal Transformer (ours) & \textbf{26.86} & 25.33 & \textbf{13.48} & \textbf{13.01} & \textbf{24.84} & 24.38\\ [1ex] 
 
 \hline
 \multicolumn{1}{|c}{ } &
 \multicolumn{6}{c|}{Rossiya Segodnya} \\ \hline
  First Sentence & 24.08 & \textbf{45.58} & 10.57 & 21.30 & 16.70 & \textbf{41.67} \\
 Encoder-Decoder & 39.10 & 38.31 & 22.13 & \textbf{21.75} & 36.34 & 36.34\\
 
 Universal Transformer w/ smoothing (ours) & 39.31 & 37.10 & 21.82 & 20.66 & 36.32 & 35.37\\ [1ex] 
 
 Universal Transformer (ours) & \textbf{39.75} & 37.62 & \textbf{22.15} & 21.04 & \textbf{36.81} & 35.91\\ [1ex] 
 \hline
\end{tabular}
\end{center}
\caption{ROUGE-1,2,L $F_1$ and recall scores, on NYT corpus and RIA corpus.}
\label{tab:results1}

\end{table}
In Tab.~\ref{tab:results1} we present results based on two corpora: the New York Times Annotated (NYT) corpus for English, and the Rossiya Segodnya (RIA) corpus for Russian. For the NYT corpus, we reached a new state of the art on ROUGE-1, ROUGE-2 and ROUGE-L $F_1$ scores. For the RIA corpus, since it has no previous art, we present results for the baselines and our model.\footnote{We are providing results from Tan et al. \cite{tan2017neural}, which were achieved using the NYT corpus. Unfortunately, the authors have not published all of their filtering criteria and seed for random sampling for this corpus, so we could not follow their setup completely. Therefore, these results are presented here for reference.} For our model we also experimented with label smoothing following \cite{Kim:2018:arXiv}.

\begin{table}[!hptb]
\centering
\begin{tabular}{|l|*{3}{>{\centering\arraybackslash}p{.15\linewidth}}|}
\hline
\backslashbox{Dataset}{User Preference} & Human  & Tie  & Machine   \\\hline
New York Times Annotated & 57.4 & 27.4 & 15.2    \\
Rossiya Segodnya & 54.4 & 30.6  & 15.0 \\
 \hline
\end{tabular}
\caption{Human evaluation results for NYT and RIA datasets.}
\label{tab:results-human-eval}

\end{table}
In our experiments, we noticed that some of the generated headlines are scored low by ROUGE metrics despite seeming reasonable, e.g. top sample in Tab.~\ref{tab:sums}. This lead us to a new series of experiments. We conducted human evaluation of obtained results for both NYT and RIA corpora. The results are presented in Tab.~\ref{tab:results-human-eval}. 5 annotators marked up 100 randomly sampled articles from a train set of each corpora. Each number shows the percentage of annotator preference over three possible options: original headline (Human), generated headline (Machine), no preference (Tie). 

For the both corpora, we could see that our model is not reaching human parity yet, having 42.6\% and 45.6\% of (Machine + Tie) user preference for NYT and RIA datasets respectively, but this result is already close to human parity and leaves  room for improvement. 

\begin{table}[!tbh]
\begin{center}
\tiny
 \begin{tabularx}{\linewidth}{|X|X|} 
 \hline



 \
 \textbf{Original text, truncated}: Unethical and irresponsible as the assertion that antidepressant medication, an excellent treatment for some forms of depression, will turn a man into a fish. It does a disservice to psychoanalysis, which offers rich and valuable insights into the human mind. ... Homosexuality is not an illness by any of the usual criteria in medicine, such as an increased risk of morbidity or mortality, painful symptoms or social, interpersonal or occupational dysfunction as a result of homosexuality itself...\\
 \textbf{Original headline}: homosexuality, not an illness, can't be cured \\
 \textbf{Generated headline}: why we can't let gay therapy begin \\
 \hline
 
  \
 \textbf{Original text, truncated}: southwest airlines said yesterday that it would add 16 flights a day from chicago midway airport, moving to protect a valuable hub amid the fight breaking out over the assets of ata airlines, the airport's biggest carrier. southwest said that beginning in january, it would add the flights to 13 cities that it already served from midway...\\
 \textbf{Original headline}: southwest is adding flights to protect its chicago hub \\
 \textbf{Generated headline}: southwest airlines to add 16 flights from chicago \\
 \hline
 
   \
 \textbf{Original text, truncated}:  москва, 1 апр - риа новости. количество сделок продажи элитных квартир в москве выросло в первом квартале этого года, по сравнению с аналогичным периодом предыдущего, в два раза, говорится в отчете компании intermarksavill s. при этом, также сообщается в нем, количество заключенных в столице первичных сделок в сегменте бизнес-класса в первом квартале 2010 года оказалось на 20 выше, чем в первом квартале прошлого года...\\
 \textbf{Original headline}: продажи элитного жилья в москве увеличились в 1 квартале в два раза\\
 \textbf{Generated headline}: продажи элитных квартир в москве в 1 квартале выросли вдвое \\
 \hline

\end{tabularx}
\end{center}

\caption{Samples of headlines generated by our model. }
\label{tab:sums}

\end{table}
\section{Conclusion}
In this paper, we explore the application of Universal Transformer architecture to the task of abstractive headline generation and outperform the abstractive state-of-the-art result on the New York Times Annotated corpus. We also present a newly released Rossiya Segodnya corpus and results achieved by our model applied to it.

\subsubsection{Acknowledgments.} Authors are thankful to Alexey Samarin for useful discussions, David Prince for proofreading, Madina Kabirova for proofreading and human evaluation organization, Anastasia Semenyuk and Maria Zaharova for help obtaining the New York Times Annotated corpus, and Alexey Filippovskii for providing the Rossiya Segodnya corpus.



\end{document}